\documentclass{amia}
\usepackage{amsmath,amssymb,amsfonts}
\usepackage{algorithmic}
\usepackage{graphicx}
\usepackage{textcomp}
\usepackage{xcolor}
\usepackage{booktabs}
\usepackage{enumitem}
\usepackage{bm}
\usepackage{multirow}
\usepackage{multicol}
\usepackage{amsfonts}
\usepackage{graphicx}
\usepackage{tabularx}
\usepackage{subcaption}
\usepackage{url}
\usepackage{calc}
\usepackage{balance}
\usepackage{float}
\usepackage{hyperref}
\usepackage{titlesec}
\titlelabel{}
\titleformat{\section}
  {\normalfont\Large\bfseries}  
  {}     %
  {0em}                         
  {}                            

\usepackage[framemethod=tikz]{mdframed}
\mdfdefinestyle{customstyle}{
  linecolor=gray!100,
  linewidth=3pt,
  innerleftmargin=3pt,
  topline=false,
  rightline=false,
  bottomline=false,
  leftline=true,
  innerrightmargin=3pt,
  innertopmargin=3pt,
  innerbottommargin=3pt,
  backgroundcolor=gray!15
}

\newenvironment{custommdframed}
  {\begin{mdframed}[style=customstyle]}
  {\end{mdframed}}

\makeatletter
\renewenvironment{abstract}{%
  \noindent\textbf{Abstract}\par\noindent
}{%
  \par
}
\makeatother



\newcommand{\unnumberedfootnote}[1]{%
    {\let\thefootnote\relax\footnotetext{#1}}%
}

\newcommand{\modelname}{\texttt{TAMER}}

\begin{document}

\title{\modelname{}: A Test-Time Adaptive MoE-Driven Framework for EHR Representation Learning}

\author{Yinghao Zhu, MS$^{1,\ast}$, Xiaochen Zheng, MS$^{1,\ast, \dagger}$, Ahmed Allam, PhD$^1$, \\Michael Krauthammer, MD, PhD$^1$}

\institutes{
    $^1$ Department of Quantitative Biomedicine, University of Zurich, Zurich, Switzerland
}

\unnumberedfootnote{$^\ast$ Equal contribution, $^\dagger$ Corresponding to Xiaochen Zheng (xiaochen.zheng@uzh.ch).}

\maketitle

\begin{abstract}
\textit{We propose \modelname{}, a \underline{T}est-time \underline{A}daptive \underline{M}oE-driven framework for Electronic Health Record (\underline{E}HR) \underline{R}epresentation learning. 
\modelname{} introduces a framework where a Mixture-of-Experts (MoE) architecture is co-designed with Test-Time Adaptation (TTA) to jointly mitigate the intertwined challenges of patient heterogeneity and distribution shifts in EHR modeling. The MoE focuses on latent patient subgroups through domain-aware expert specialization, while TTA enables real-time adaptation to evolving health status distributions when new patient samples are introduced. Extensive experiments across four real-world EHR datasets demonstrate that \modelname{} consistently improves predictive performance for both mortality and readmission risk tasks when combined with diverse EHR modeling backbones. \modelname{} offers a promising approach for dynamic and personalized EHR-based predictions in practical clinical settings.}
\end{abstract}

\section{Introduction}

Electronic Health Records (EHRs) have revolutionized healthcare data management and analysis, offering unprecedented opportunities for personalized medicine~\cite{gao2024comprehensive,chowdhury2023primenet,zhu2024prism,ma2020concare}. However, in real-world deployment, EHR-based predictive models face two primary challenges: first, the inherent heterogeneity of patient populations~\cite{heterogeneity1,heterogeneity2,heterogeneity3} necessitates adaptive and personalized approaches to capture the diverse health status of patient subtypes or subgroups~\cite{huo2021sparse,fu2020assessment}. Second, the dynamic nature of clinical data, where new patient information arrives continuously and often without immediate outcome labels, requires models to adapt to changing health status distributions and learn from unlabeled data in real-time~\cite{bellandi2024data}.

To address the challenge of patient heterogeneity, we propose and build on the Mixture-of-Experts (MoE) modeling paradigm inspired by its latest application in large language models (LLMs). MoE aims to activate different experts for diverse inputs~\cite{dai2024deepseekmoe,lin2024moe} -- each expert specializes in what they excel at. This concept aligns well with healthcare practice, where specialists focus on specific conditions or areas of medicine~\cite{de2004quality,harrold1999knowledge}. 

However, simply adopting the MoE architecture does not fully address the complexities of healthcare data, particularly the dynamic nature of clinical information. In real-world applications, the distribution of patient data can shift over time and cross hospitals. As proposed by~\cite{ttt1,ttt2,ttt3,ttt4,ttaood}, self-supervised training and test-time training can reduce the distribution shift performance gap for several tasks or scenarios. To tackle this issue, we incorporate a test-time training approach, which allows the model to adapt to evolving patient data distributions. Specifically, we introduce a Test-Time Adaptation (TTA) module~\cite{sun2024learning,liang2024comprehensive} to adjust part of the network (i.e model parameters) and learn the evolving patient health status in a self-supervised manner, given that outcome labels are unavailable for new patients to update the entire neural network. This is achieved by reconstructing the hidden representation and computing the gradients when processing the new patient samples that are used to update the TTA module. By integrating TTA with MoE, our proposed \modelname{} continuously adapts patients' representations and assigns them to appropriate experts.

Existing works utilizing MoE to model EHR heterogeneity~\cite{huo2021sparse,lee2022learning,hurley2023clinical,han2024fusemoe} generally employ sparse MoE structures with top-K gating~\cite{fedus2022switch}, which may lead to representation collapse~\cite{chi2022representation} and imbalanced expert loads~\cite{zhou2022mixture}. Additionally, previous works do not consider the evolving nature of EHR data, hindering their application in online clinical settings~\cite{ttt1}. Instead, our model employs a dynamic soft MoE to overcome the representation collapse~\cite{puigcerver2023sparse}, and a TTA module to partially adjust the model's weights during inference time when processing new patient samples.

Overall, \modelname{} offers the following contributions:
\begin{enumerate}
    \item \textit{Methodologically}, \modelname{} introduces a plug-in framework that can be integrated into existing EHR feature extractors. Leveraging the MoE integrated with TTA, \modelname{} offers a unified solution to tackle both the heterogeneity of patient populations and the evolving nature of EHR data, which are crucial challenges in real-world clinical applications.
    \item \textit{Experimentally}, \modelname{} consistently improves performance across four real-world EHR datasets for both mortality and readmission prediction tasks on various EHR modeling backbones. Additional ablation studies and analyses provide insights into the ability of \modelname{} for robust performance under distribution shifts.
\end{enumerate}

\section{Problem Definition}

EHR datasets are structured as multivariate time-series data, represented by $\bm{X} = [\bm{x}_1, \bm{x}_2, \cdots, \bm{x}_T]^{\top} \in \mathbb{R}^{T \times F}$. This matrix encapsulates information fro $T$ visits across $F$ features, comprising both static (e.g., sex and age) and dynamic (e.g., lab tests and vital signs) features.

We focus on two EHR-specific binary classification tasks: in-hospital mortality and 30-day readmission prediction. Our goal is to develop \modelname{}, a plug-in model enhancing existing EHR models' feature extractors. The predictive objective is:
\begin{equation}
    \hat{y} = \texttt{Backbone}_{+\modelname{}}(\bm{X})
\end{equation}
where $\hat{y}$ is the predicted outcome. For mortality prediction, we use the initial 48-hour window of an ICU stay to predict discharge status (0: alive, 1: deceased). For readmission prediction, we forecast 30-day readmission (0: no, 1: yes).

\section{Methodology}

\begin{figure*}[!ht]
    \centering
    \includegraphics[width=1.0\linewidth]{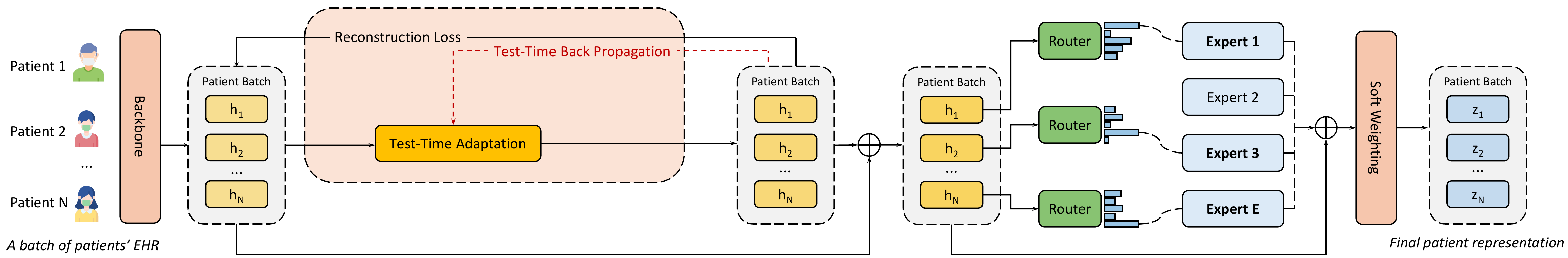}
    \caption{\textit{Overall framework of \modelname{}.}}
    \label{fig:pipeline}
\end{figure*}

As shown in Fig.~\ref{fig:pipeline}, for a batch of $N$ patients, we employ a backbone model $\bm{\theta}_m$ to generate a sequence of hidden representations for each patient $i$, expressed as $h_i = \texttt{Backbone}(x_1, ..., x_T)$. The hidden state $h_i$ encapsulates the patient's historical information up to their last visit.

We propose integrating a Test-Time Adaptation (TTA) layer between the backbone model and the Mixture-of-Experts (MoE) module. This TTA layer allows the model to adapt its parameters during test time through a self-supervised approach~\cite{ttt1}. In this work, we focus on a reconstruction task, where the goal is to reconstruct the hidden representations $h_i$ from the backbone model using the output of the TTA layer. Specifically, the TTA layer is implemented as a Multi-Layer Perceptron (MLP) with weights $\bm{\theta}_s$, which are updated dynamically for each test instance based on the Mean Squared Error (MSE) loss, denoted as $l_s$. The central idea behind TTA is to enable the model to adapt its parameters in response to the specific test sample, thereby creating a more flexible and adaptive decision boundary.

Following the TTA layer, we implement a Mixture-of-Experts (MoE) module, parameterized by $\bm{\theta}_e$. This module consists of multiple expert networks, each specializing in different aspects of the EHR data. A softmax-based router assigns each patient to different experts, allowing for a more nuanced and comprehensive patient representation. Within the MoE module, a gating network assigns weights to each expert's output. The final patient representation $z_i$ is formed by a soft-weighted combination of these expert outputs, as described by~\cite{puigcerver2023sparse}. This approach integrates specialized knowledge from multiple domains, aiming to deliver a more robust and adaptable model for patient analysis.

Note that residual connections~\cite{he2016deep} are employed in both the MoE and TTA components for improved gradient flow.

\textbf{\raisebox{0.1em}{$\bullet$}\hspace{0.5em}Train-time Training.} Our main training objective is to minimize the empirical risk:
\begin{equation}
    \label{lm}
    \underset{\bm{\theta}_m, \bm{\theta}_s, \bm{\theta}_e}{\text{min}}\frac{1}{n} \sum_{i=1}^n \bigl[l_m (\bm{X}_i, y_i;\bm{\theta}_m,\bm{\theta}_e) + l_s (\texttt{Backbone}(\bm{X}_i);\bm{\theta}_s) \bigr]
\end{equation}

We compute the summation loss for $n$ samples across the training set using a supervised learning paradigm. The main target loss $l_m$ is defined as the binary cross entropy loss for binary classification tasks in our scenario, while we simultaneously minimize the auxiliary MSE loss $l_s$ on $\texttt{Backbone}(\bm{X})$ for TTA layer.

\textbf{\raisebox{0.1em}{$\bullet$}\hspace{0.5em}Test-time Training.}
Instead of only optimizing model parameters in the training stage, \modelname{} updates the TTA layer's parameters across all phases (training, validation, and test). Specifically, in validation and test stage, we freeze the backbone and the MoE module. The objective is:
\begin{align}
    \label{ls}
    \underset{\bm{\theta}_s}{\text{min}}\ l_s (\texttt{Backbone}(\bm{X});\bm{\theta}_s)    
\end{align}

It is important to note that we continuously update $\bm{\theta}_s$ as test samples arrive sequentially in an online manner~\cite{ttt1}.

\begin{figure}[!ht]
    \centering
    \includegraphics[width=0.7\linewidth]{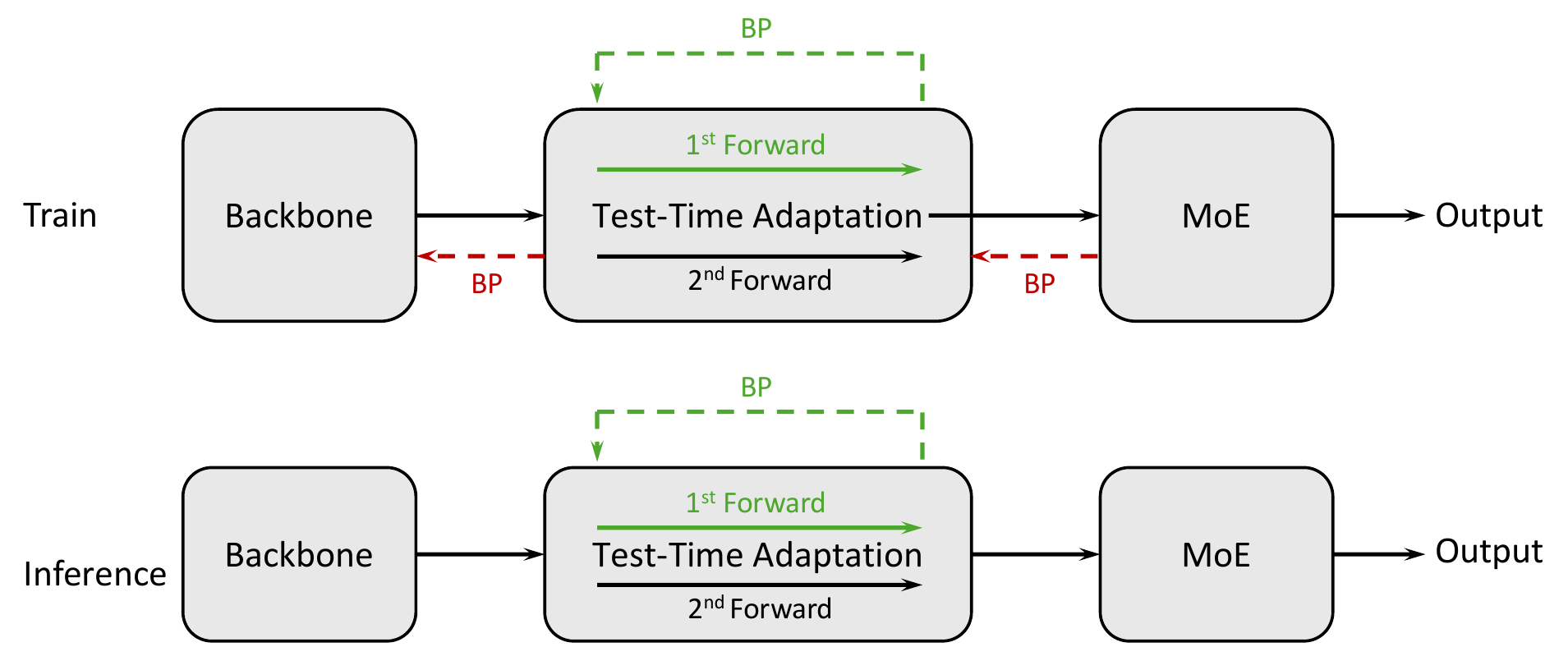}
    \caption{\textit{Computational flow of \modelname{}.} Solid line: forward pass. Dashed line: backpropagation (BP) and parameter updating.}
    \label{fig:flow}
\end{figure}

\textbf{\raisebox{0.1em}{$\bullet$}\hspace{0.5em}Computational Flow.}
As illustrated in Fig.~\ref{fig:flow}, input samples are sequentially processed through the backbone, TTA, and MoE before the loss computation or outcome prediction. The TTA parameters are updated via backpropagation after computing the reconstruction loss $l_s$. A second forward pass is done, where the generated backbone hidden states are computed using the updated TTA, producing new representations adapted to the current distribution.

\section{Experimental Setups}

\textbf{\raisebox{0.1em}{$\bullet$}\hspace{0.5em}Benchmarked Real-world Datasets.}
We evaluate our approach on four established and widely used datasets for benchmarking: MIMIC-III~\cite{johnson2016mimic}, MIMIC-IV~\cite{johnson2023mimic}, PhysioNet Challenge 2012~\cite{silva2012challenge}, and eICU~\cite{pollard2018eicu}. Each dataset is split into training (70\%), validation (10\%), and test (20\%) sets using stratified shuffle split based on patients' end-stage mortality outcomes. We use the Last Observation Carried Forward (LOCF) imputation method~\cite{woolley2009last} by default. Tab.~\ref{tab:statistics_datasets} presents the statistics of the preprocessed datasets.

\begin{table}[!ht]
    \fontsize{10}{10}\selectfont
    \centering
    \caption{\textit{Statistics of datasets after preprocessing.} Proportions indicate the percentage of labels with value $1$. ``Out.'' denotes ``mortality outcome'', ``Re.'' denotes ``Readmission.''}
    \label{tab:statistics_datasets}
    % \resizebox{0.6\linewidth}{!}{
    \begin{tabular}{c|c|c|c|c}
        \toprule
        \textbf{Dataset}      & \textbf{MIMIC-III} & \textbf{MIMIC-IV} & \textbf{Challenge-2012} & \textbf{eICU}    \\
        \midrule
        \# Samples            & 41517     & 56888    & 4000           & 73386   \\
        % Missing               & 69.87\%   & 74.70\%  & 84.68\%        & 42.61\% \\
        $\text{Label}_{\text{Out.}}$ & 10.62\%   & 9.55\%   & 13.85\%        & 8.32\%  \\
        $\text{Label}_{\text{Re.}}$  & 14.74\%   & 13.85\%  & -              & -       \\
        \# Static & 2 & 2 & 5 & 2\\
        \# Dynamic & 59 & 59 & 35 & 12\\
        \bottomrule
    \end{tabular}
    % }
\end{table}

\textbf{\raisebox{0.1em}{$\bullet$}\hspace{0.5em}Evaluation Metrics.}
We evaluate binary classification performance using AUROC and AUPRC. We emphasize AUPRC as the primary metric due to its informativeness when dealing with highly imbalanced and skewed datasets~\cite{kim2022auprc} (such as the four datasets we are using). All reported results include mean±std computed from 100 bootstrap samples of the test set.

\textbf{\raisebox{0.1em}{$\bullet$}\hspace{0.5em}Baseline Models.}
We include conventional sequential modeling models: GRU and Transformer, EHR-specific models: RETAIN~\cite{choi2016retain}, AdaCare~\cite{ma2020adacare}, ConCare~\cite{ma2020concare}, GRASP~\cite{zhang2021grasp}, M3Care~\cite{zhang2022m3care}, AICare~\cite{ma2023aicare}, and PrimeNet~\cite{chowdhury2023primenet}. Each model is compared with \modelname{}, which serves as the plug-in module.

\textbf{\raisebox{0.1em}{$\bullet$}\hspace{0.5em}Implementation Details.}
All experiments are conducted on a single Nvidia RTX 3090 GPU with CUDA 12.5. The server's system memory (RAM) size is 128GB. We implement the model in Python 3.9.19, PyTorch 2.3.0~\cite{paszke2019pytorch}, PyTorch Lightning 2.2.4~\cite{falcon2019lightning}, and pyehr~\cite{zhu2024pyehr}. We employ AdamW~\cite{loshchilov2017decoupled} with a batch size of 1024 patients. Models are trained for 100 epochs with early stopping based on AUPRC after 10 epochs without improvement. Learning rates $\{0.01, 0.001, 0.0001\}$ and hidden dimensions $\{64, 128\}$ are tuned using grid search on the validation set. Random seeds are set to 0. Performance is reported in the form of mean±std of bootstrapping on all test set samples 100 times.

We adapt the code implementations of TTA~\cite{sun2024learning} and soft MoE~\cite{puigcerver2023sparse} from their respective GitHub repositories\footnote{TTA: \url{https://github.com/kyegomez/TTL}}\textsuperscript{,}\footnote{Soft MoE: \url{https://github.com/lucidrains/soft-moe-pytorch}}.

\section{Experimental Results and Analysis}

To evaluate and analyze the effectiveness of \modelname{}, we propose to answer the following research questions (RQs).

\begin{itemize}
    \item \textbf{RQ1: Overall In-domain Experimental Results.} Can \modelname{} serve as a plug-in module for diverse backbone models and enhance their predictive performance?
    \item \textbf{RQ2: Performance under Distributional Shift Conditions.} How effectively does \modelname{} adapt to distribution shifts in EHR?
    \item \textbf{RQ3: Ablation Study.} What is the role of the proposed modules (MoE module and Test-Time Adaptation (TTA) module) in the model's performance?
    \item \textbf{RQ4: Sensitivity to Number of Experts.} What is the optimal number of experts? How does the number of experts affect the model's performance?
    \item \textbf{RQ5: Sensitivity to LR in TTA.} What is the best learning rate (LR) to use in the TTA module?
    \item \textbf{RQ6: Parameter Size and Computational Complexity Analysis.} How does incremental parameter augmentation affect the computational efficiency and predictive performance of base models?
\end{itemize}

\textbf{\raisebox{0.1em}{$\bullet$}\hspace{0.5em} RQ1: Overall In-domain Experimental Results.}\label{sec:Experimental_Results}
To address RQ1, we first compare performance with and without \modelname{} on various EHR modeling backbone models, which serve as feature extractors. We integrate the \modelname{} architecture after each backbone. Results in Tab.~\ref{tab:experimental_results} demonstrate \modelname{}'s consistent effectiveness across datasets and model backbones. Specifically, \modelname{} integration improves performance in $\frac{53}{54}$ instances according to the AUPRC metric. The results of the paired t-tests confirm that the performance improvements achieved by \modelname{} are statistically significant ($p < 0.05$) across all evaluated datasets and baseline models.

\begin{table*}[!t]
        \centering
        \caption{\textit{Benchmarking results on MIMIC-III, MIMIC-IV, Challenge-2012, and eICU datasets for mortality prediction and 30-day readmission tasks.} We compare performance with and without \modelname{} on diverse EHR backbone models. \textbf{Bold} indicates that the model with \modelname{} has better or equal performance compared to the baseline.}
        \label{tab:experimental_results}
\resizebox{\textwidth}{!}{
    \begin{tabular}{c|cc|cc|cc|cc|cc|cc}
\toprule
\textbf{Dataset}     & \multicolumn{2}{c|}{\textbf{MIMIC-III Mortality}} & \multicolumn{2}{c|}{\textbf{MIMIC-III Readmission}} & \multicolumn{2}{c|}{\textbf{MIMIC-IV Mortality}} & \multicolumn{2}{c|}{\textbf{MIMIC-IV Readmission}} & \multicolumn{2}{c|}{\textbf{Challenge-2012 Mortality}} & \multicolumn{2}{c}{\textbf{eICU Mortality}} \\ \midrule
\textbf{Metric}      & AUPRC($\uparrow$) & AUROC($\uparrow$) & AUPRC($\uparrow$)   & AUROC($\uparrow$) & AUPRC($\uparrow$)   & AUROC($\uparrow$)  & AUPRC($\uparrow$)    & AUROC($\uparrow$)   & AUPRC($\uparrow$)  & AUROC($\uparrow$)  & AUPRC($\uparrow$)   & AUROC($\uparrow$)  \\ \midrule

GRU & {52.96±2.24} & {85.09±0.87}  & 52.76±2.30 & 85.66±0.75 & {50.68±2.03} & {84.71±0.80} & 50.07±2.08 & 84.75±0.85 & {34.65±4.00} & {74.72±1.92} & 45.76±1.48 & 85.48±0.57 \\
\hspace{2mm}+\modelname{} & \textbf{54.63±2.44} & \textbf{86.48±0.79} & \textbf{54.33±2.29} & \textbf{85.91±0.77} & \textbf{56.33±1.93} & \textbf{86.64±0.76} & \textbf{54.73±2.03} & \textbf{86.24±0.81} & \textbf{38.95±4.37} & \textbf{78.70±2.09} & \textbf{47.12±1.61} & \textbf{85.92±0.54} \\
\midrule

Transformer & 43.37±2.39 & 82.14±1.00 & 44.78±2.42 & 81.88±1.00 & 43.52±1.85 & 83.58±0.84 & 41.65±1.78 & 83.50±0.86 & {36.33±4.00} & {78.86±1.81} & 35.67±1.37 & 79.32±0.61 \\
\hspace{2mm}+\modelname{} & \textbf{49.15±2.40} & \textbf{84.79±0.91} & \textbf{46.26±2.41} & \textbf{83.31±0.90} & \textbf{46.64±1.91} & \textbf{84.38±0.87} & \textbf{42.25±1.80} & 83.38±0.84 & \textbf{39.15±4.14} & 78.67±1.82 & \textbf{36.19±1.39} & 79.10±0.67 \\
\midrule

RETAIN & 50.30±2.35 & 85.52±0.81 & 48.73±2.33 & 84.25±0.86 & 53.31±2.14 & 86.09±0.79 & 46.84±1.94 & 84.30±0.82 & {37.28±3.85} & {78.51±1.76} & 41.16±1.52 & 81.19±0.62  \\
\hspace{2mm}+\modelname{} & \textbf{50.60±2.34} & \textbf{85.52±0.81} & \textbf{49.28±2.50} & 82.60±1.02 & \textbf{54.82±2.06} & \textbf{86.36±0.75} & \textbf{52.45±2.02} & \textbf{86.17±0.78} & \textbf{39.02±4.52} & 78.19±2.14 & \textbf{43.38±1.52} & \textbf{83.60±0.59}  \\
\midrule

AdaCare & 52.25±2.34 & 85.29±0.80 & 51.30±2.28 & 84.33±0.84 & 48.51±2.00 & 83.54±0.92& 50.26±2.08 & 84.29±0.87 & 43.75±4.80 & 78.54±2.10 & 45.34±1.48 & 84.15±0.55  \\
\hspace{2mm}+\modelname{} & \textbf{52.99±2.27} & \textbf{85.56±0.77} & \textbf{51.66±2.27} & \textbf{85.41±0.77} & \textbf{50.95±2.00} & \textbf{84.44±0.85} & \textbf{50.73±2.12} & 83.99±0.96 & \textbf{44.53±4.64} & \textbf{79.39±1.84} & \textbf{45.65±1.48} & \textbf{84.50±0.54}  \\
\midrule

ConCare & 54.75±2.35 & 86.73±0.76 & 51.91±2.44 & 85.74±0.78 & 53.81±2.05 & 85.54±0.83 & 51.01±2.06 & 84.63±0.84 & {49.30±4.28} & {83.31±1.91} & 48.00±1.40 & 85.41±0.52  \\
\hspace{2mm}+\modelname{} & \textbf{56.17±2.24} & \textbf{86.94±0.77} & \textbf{52.79±2.43} & \textbf{86.04±0.80} & 53.76±1.97 & \textbf{85.71±0.83} & \textbf{52.73±2.03} & \textbf{84.83±0.90} & \textbf{51.57±4.68} & \textbf{85.22±1.78} & \textbf{48.01±1.41} & \textbf{85.57±0.51} \\
\midrule

GRASP & 54.43±2.29 & 85.98±0.81 & 53.14±2.29 & 85.40±0.79 & 53.72±2.00 & 85.55±0.82 & 54.38±2.04 & 86.19±0.77 & 40.33±4.10 & 77.28±2.02 & 45.52±1.47 & 83.90±0.61 \\
\hspace{2mm}+\modelname{} & \textbf{54.94±2.36} & \textbf{86.67±0.81} & \textbf{54.36±2.29} & \textbf{85.63±0.77} & \textbf{53.95±1.96} & \textbf{85.74±0.79} & \textbf{54.60±2.06} & 86.16±0.77 & \textbf{44.14±4.24} & \textbf{82.11±1.93} & \textbf{46.75±1.47} & \textbf{84.21±0.57}  \\
\midrule

M3Care & 53.21±2.40 & 86.50±0.73 & 50.27±2.34 & 84.74±0.79 & 53.09±1.92 & 84.56±0.87 & 53.15±1.86 & 85.29±0.83 & {31.36±4.08} & {68.50±2.73} & 45.26±1.45 & 83.71±0.56  \\
\hspace{2mm}+\modelname{} & \textbf{53.24±2.36} & \textbf{86.70±0.72} & \textbf{51.84±2.52} & \textbf{86.04±0.75} & \textbf{53.88±1.93} & \textbf{85.07±0.85} & \textbf{53.93±1.85} & \textbf{85.69±0.80} & \textbf{33.40±4.35} & \textbf{73.30±2.62} & \textbf{45.87±1.43} & \textbf{84.41±0.53}  \\
\midrule

AICare & 49.27±2.24 & 84.15±0.87 & 44.70±2.37 & 82.14±1.01 & 49.27±2.24 & 84.15±0.87 & 46.93±1.92 & 83.79±0.82 & {47.19±4.47} & {85.95±1.53} & 43.02±1.51 & 83.17±0.54 \\
\hspace{2mm}+\modelname{} & \textbf{49.38±2.63} & \textbf{84.34±0.93} & \textbf{47.11±2.43} & \textbf{83.73±0.90} & \textbf{50.28±2.03} & \textbf{84.94±0.79} & \textbf{48.72±2.00} & \textbf{83.79±0.85} & \textbf{47.35±4.57} & 83.12±1.92 & \textbf{43.28±1.56} & 82.73±0.58\\
\midrule

PrimeNet & 46.06±2.38 & 83.78±0.81 & 44.65±2.53 & 82.57±1.00 & 47.43±2.01 & 84.11±0.86 & 46.17±2.12 & 83.87±0.89 & 36.33±4.12 & 75.85±2.08 & 43.73±1.51 & 83.10±0.59  \\
\hspace{2mm}+\modelname{} & \textbf{49.11±2.58} & \textbf{85.02±0.83} & \textbf{46.18±2.50} & 82.15±0.89 & \textbf{48.00±2.17} & \textbf{84.19±0.87} & \textbf{47.61±2.12} & \textbf{84.25±0.89} & \textbf{36.59±4.20} & 72.97±2.34 & \textbf{44.52±1.43} & \textbf{84.13±0.54} \\ 
\bottomrule
\end{tabular}
}
\end{table*}

Although \modelname{} is designed as a plug-in module, we conduct comprehensive comparisons with state-of-the-art MoE methods in the EHR domain and test-time training approaches to demonstrate its performance enhancements. As shown in Tab.~\ref{tab:baseline_comparison}, \modelname{} demonstrates better performance across both evaluation metrics. Among MoE-based methods, Soft MoE achieves more competitive AUPRC compared to other sparse MoE implementations, likely because sparse assignments can lead to representation collapse~\cite{chi2022representation} and imbalanced expert loads~\cite{zhou2022mixture}. In the test-time training comparisons, \modelname{} shows a 1.35\% relative improvement over TTT in AUPRC while maintaining better AUROC performance. These results validate that our integrated approach effectively leverages both patient heterogeneity modeling through Soft MoE and distribution adaptation through TTA, resulting in more robust performance compared to methods that do not account for these EHR-specific characteristics.

\begin{table}[!ht]
    \vspace{10pt}
\fontsize{10}{10}\selectfont
    \centering
    \caption{\textit{Comparison with MoE and test-time training baselines on MIMIC-IV mortality prediction.}}
    \label{tab:baseline_comparison}
    % \resizebox{0.6\linewidth}{!}{
    \begin{tabular}{c|c|c|c}
        \toprule
        \textbf{Type} & \textbf{Model} & \textbf{AUPRC}($\uparrow$) & \textbf{AUROC}($\uparrow$) \\
        \midrule
        \multirow{3}{*}{MoE} & Soft MoE~\cite{puigcerver2023sparse} & {55.93±1.85} & {86.45±0.79}  \\
        & Sparse Gated MoE~\cite{huo2021sparse} & {53.73±2.01} & {86.09±0.80}  \\
        & Outcomes-driven MoE~\cite{hurley2023clinical} & {55.18±1.91} & {86.50±0.78} \\
        \midrule
        \multirow{1}{*}{Test-time} & TTT~\cite{sun2024learning} & {55.58±1.91} & {86.22±0.79} \\
        \midrule
        Integrated & \modelname{} & \textbf{56.33±1.93} & \textbf{86.64±0.76} \\
        \bottomrule
    \end{tabular}
    % }
\end{table}

\begin{custommdframed}
\textit{Answer to RQ1 (Overall In-domain Experimental Results):} \modelname{} could serve as an effective plug-in module for enhancing predictive performance in EHR-based models.
\end{custommdframed}

\textbf{\raisebox{0.1em}{$\bullet$}\hspace{0.5em}RQ2: Performance under Distributional Shift Conditions.}\label{sec:Distribution_Shift}
To address RQ2, we conduct cross-dataset experiments using MIMIC-III and MIMIC-IV to evaluate our model’s ability to handle shifts in patient data distribution over time and across different hospital systems. As reported in Tab.~\ref{tab:cross_dataset}, \modelname{}—equipped with the TTA module—consistently achieves higher performance metrics compared to baseline methods that lack this adaptation mechanism. Specifically, \modelname{} trained on MIMIC-III demonstrates noticeable generalization when tested on MIMIC-IV, and vice versa, indicating that our TTA module mitigates the performance degradation caused by distributional discrepancies. These findings suggest that \modelname{} captures and adjusts to variations in patient populations and data collection procedures across different periods and hospital systems, thereby validating the efficacy of the TTA module in real-world clinical settings.

\begin{table}[ht]
    \fontsize{10}{10}\selectfont
    \centering
    \caption{\textit{Cross-dataset evaluation results for mortality prediction between MIMIC-III and MIMIC-IV.} \textbf{Bold} indicates that the model with \modelname{} has better or equal performance compared to the baseline.}
    \label{tab:cross_dataset}
    % \resizebox{0.6\linewidth}{!}{
    \begin{tabular}{c|cc|cc}
        \toprule
        \multirow{2}{*}{\textbf{Model}} & \multicolumn{2}{c|}{\textbf{MIMIC-III→IV}} & \multicolumn{2}{c}{\textbf{MIMIC-IV→III}} \\
        \cmidrule(lr){2-3} \cmidrule(lr){4-5}
        & AUPRC($\uparrow$) & AUROC($\uparrow$) & AUPRC($\uparrow$) & AUROC($\uparrow$) \\
        \midrule
        GRU & 48.37±0.98 & 81.93±0.34 & 31.17±0.89 & 73.45±0.49 \\
        \hspace{2mm}+\modelname{} & \textbf{51.84±1.08} & \textbf{84.28±0.34} & \textbf{31.47±0.91} & \textbf{75.48±0.46} \\
        \midrule
        Transformer & 43.76±0.97 & 83.33±0.32 & 30.65±0.82 & 75.74±0.46 \\
        \hspace{2mm}+\modelname{} & \textbf{44.84±1.00} & \textbf{83.71±0.32} & \textbf{31.17±0.82} & \textbf{76.21±0.46} \\
        \midrule
        RETAIN & 48.30±1.07 & 84.04±0.38 & 30.18±1.04 & 68.58±0.66 \\
        \hspace{2mm}+\modelname{} & \textbf{48.44±1.15} & \textbf{84.44±0.37} & \textbf{30.51±0.88} & \textbf{70.33±0.59} \\
        \midrule
        AdaCare & 51.76±0.95 & 85.30±0.36 & 26.79±0.81 & 72.01±0.51 \\
        \hspace{2mm}+\modelname{} & \textbf{52.66±0.95} & \textbf{85.53±0.35} & \textbf{28.75±0.84} & \textbf{74.21±0.49} \\
        \midrule
        ConCare & 51.77±1.02 & 85.67±0.36 & 31.85±0.86 & 77.57±0.45 \\
        \hspace{2mm}+\modelname{} & \textbf{52.50±0.96} & \textbf{85.80±0.33} & 30.53±0.83 & 76.37±0.47 \\
        \midrule
        GRASP & 53.16±0.98 & 85.70±0.34 & 31.57±0.91 & 73.21±0.48 \\
        \hspace{2mm}+\modelname{} & \textbf{54.23±0.99} & \textbf{86.23±0.35} & \textbf{38.08±0.98} & \textbf{74.74±0.50} \\
        \midrule
        M3Care & 49.86±0.92 & 84.43±0.34 & 31.56±0.94 & 75.37±0.50 \\
        \hspace{2mm}+\modelname{} & \textbf{51.80±0.94} & \textbf{85.19±0.34} & \textbf{34.71±0.97} & \textbf{76.64±0.47} \\
        \midrule
        AICare & 48.35±1.02 & 84.65±0.34 & 29.88±0.86 & 74.29±0.48 \\
        \hspace{2mm}+\modelname{} & \textbf{49.58±0.93} & \textbf{84.74±0.35} & \textbf{37.00±1.05} & \textbf{75.94±0.45} \\
        \midrule
        PrimeNet & 47.43±1.02 & 83.45±0.33 & 27.74±0.76 & 74.09±0.49 \\
        \hspace{2mm}+\modelname{} & \textbf{48.26±1.03} & \textbf{83.97±0.34} & \textbf{28.80±0.78} & \textbf{75.39±0.49} \\
        \bottomrule
    \end{tabular}
    % }
\end{table}

\begin{custommdframed}
\textit{Answer to RQ2 (Performance under Distributional Shift Conditions):} \modelname{} effectively handles data distribution changes between different hospital systems, as shown by its performance in cross-dataset testing between MIMIC-III and MIMIC-IV.
\end{custommdframed}

\textbf{\raisebox{0.1em}{$\bullet$}\hspace{0.5em}RQ3: Ablation Study.} \label{sec:Ablation_Study}
We conduct an ablation study by comparing reduced versions of \modelname{}: removing MoE, TTA, and altering TTA placement relative to MoE. Results in Tab.~\ref{tab:ablation_study} demonstrate that both MoE and TTA modules contribute significantly to \modelname{}'s performance. The combination of MoE and TTA (placed before MoE) yields the best results across both datasets. Notably, placing TTA after MoE worsens performance. This aligns with intuition, as adapting the distribution before expert assignment is more logical than after.

\begin{table}[H]
    \vspace{10pt}
    \fontsize{10}{10}\selectfont
    \centering
    \caption{\textit{Ablation study of \modelname{} components on MIMIC-III and MIMIC-IV mortality prediction tasks.} TTA module's selection denotes the placement relative to MoE. The backbone is GRU model.}
    \label{tab:ablation_study}
    % \resizebox{0.6\linewidth}{!}{
    \begin{tabular}{ccccccc}
        \toprule
        \multirow{2}{*}{\textbf{MoE}} & \multicolumn{2}{c}{\textbf{TTA}} & \multicolumn{2}{c}{\textbf{MIMIC-III Mortality}} & \multicolumn{2}{c}{\textbf{MIMIC-IV Mortality}} \\
        \cmidrule(lr){2-3} \cmidrule(lr){4-5} \cmidrule(lr){6-7}
        & Before  & After & AUPRC & AUROC & AUPRC & AUROC \\
        \midrule
        - & - & - & {52.96±2.24} & {85.09±0.87} & {50.68±2.03} & {84.71±0.80} \\
        - & \checkmark & - & {54.50±2.21} & {86.12±0.80} & {55.58±1.91} & {86.22±0.79} \\
        \checkmark & - & - & {53.62±2.33} & {86.37±0.81} & {55.93±1.85} & {86.45±0.79} \\
        \checkmark & \checkmark & - & \textbf{54.63±2.44} & \textbf{86.48±0.79} & \textbf{56.33±1.93} & \textbf{86.64±0.76} \\
        \checkmark & - & \checkmark & {52.80±2.31} & {86.49±0.74} & {53.19±1.98} & {85.19±0.82} \\
        \bottomrule
    \end{tabular}
    % }
\end{table}

\begin{custommdframed}
\textit{Answer to RQ3 (Ablation study):} Both MoE and TTA modules are effective in EHR modeling; in addition, it is better to place TTA before MoE.
\end{custommdframed}

\textbf{\raisebox{0.1em}{$\bullet$}\hspace{0.5em}RQ4: Sensitivity to number of experts.}
To investigate the effect of the number of experts in the Mixture-of-Experts (MoE) module, we vary the number of experts (2, 4, 8, 16, and 32). Results in Tab.~\ref{tab:num_moe_experts} reveal a complex relationship between expert count and performance: the performance fluctuates non-monotonically as the expert count increases. While 16 experts perform best for both MIMIC-III and MIMIC-IV datasets, the optimal count likely varies depending on the dataset properties.

\begin{table}[H]
    \fontsize{10}{10}\selectfont
    \centering
    \caption{\textit{Effect of the number of experts in MoE on \modelname{} performance for the MIMIC-III and MIMIC-IV mortality prediction task.} The backbone is GRU model.}
    \label{tab:num_moe_experts}
    % \resizebox{0.6\linewidth}{!}{
    \begin{tabular}{ccccc}
        \toprule
        \multirow{2}{*}{\textbf{\# Experts}} & \multicolumn{2}{c}{\textbf{MIMIC-III}} & \multicolumn{2}{c}{\textbf{MIMIC-IV}} \\
        \cmidrule(lr){2-3} \cmidrule(lr){4-5}
        & AUPRC & AUROC & AUPRC & AUROC \\
        \midrule
        2 & {53.85±2.47} & {86.11±0.86} & {53.75±2.05} & {85.95±0.80} \\
        4 & {53.37±2.33} & {85.32±0.87} & {53.85±1.96} & {85.12±0.84} \\
        8 & {53.09±2.33} & {86.32±0.76} & {51.88±2.11} & {84.97±0.86} \\
        16 & \textbf{54.63±2.44} & \textbf{86.48±0.79} & \textbf{56.33±1.93} & \textbf{86.64±0.76} \\
        32 & {52.60±2.34} & {85.31±0.84} & {54.02±2.14} & {85.71±0.86} \\
        \bottomrule
    \end{tabular}
    % }
\end{table}

\begin{custommdframed}
\textit{Answer to RQ4 (Sensitivity to number of experts):} The optimal number of experts for both MIMIC datasets is 16. However, this optimal number may vary depending on the size and complexity of the datasets under consideration.
\end{custommdframed}

\textbf{\raisebox{0.1em}{$\bullet$}\hspace{0.5em}RQ5: Sensitivity to LR in TTA.}
We explore the sensitivity of \modelname{} to different learning rates (LR) in the TTA module for gradient updates during test time. The results in Tab.~\ref{tab:lr_tta} indicate that a model with the TTA module generally outperforms a model without (see first row denoted by None, which can be treated as a GRU model connected with MoE only). A learning rate of 0.00001 yields the best performance for both datasets. Notably, with a large LR (0.1), the performance is worse than with no gradient updates, possibly due to excessively large update steps resulting in suboptimal updated weights.

\begin{table}[H]
    \vspace{10pt}
    \fontsize{10}{10}\selectfont
    \centering
    \caption{\textit{Effect of different learning rates (LR) in TTA module for gradient updates for the MIMIC-III and MIMIC-IV mortality prediction task.} None means no gradient updates. The backbone is GRU model.}
    \label{tab:lr_tta}
    % \resizebox{0.5\linewidth}{!}{
    \begin{tabular}{ccccc}
        \toprule
        \multirow{2}{*}{\textbf{LR}} & \multicolumn{2}{c}{\textbf{MIMIC-III}} & \multicolumn{2}{c}{\textbf{MIMIC-IV}} \\
        \cmidrule(lr){2-3} \cmidrule(lr){4-5}
        & AUPRC & AUROC & AUPRC & AUROC \\
        \midrule
        None & {53.32±2.30} & {86.34±0.78} & {52.67±1.99} & {85.16±0.83} \\
        0.00001 & \textbf{54.63±2.44} & \textbf{86.48±0.79} & \textbf{56.33±1.93} & \textbf{86.64±0.76} \\
        0.0001 & {54.09±2.15} & {85.54±0.81} & {53.57±1.92} & {85.82±0.79} \\
        0.001 & {52.40±2.27} & {84.85±0.81} & {52.80±2.08} & {85.32±0.88} \\
        0.01 & {53.44±2.34} & {85.50±0.83} & {51.02±2.09} & {84.31±0.87} \\
        0.1 & {51.74±2.14} & {85.03±0.77} & {52.49±1.95} & {85.40±0.80} \\
        \bottomrule
    \end{tabular}
    % }
\end{table}

\begin{custommdframed}
\textit{Answer to RQ5 (Sensitivity to LR in TTA):} Performance generally decreases with higher learning rates, suggesting that small and gradual updates are more effective for adaptation.
\end{custommdframed}

\textbf{\raisebox{0.1em}{$\bullet$}\hspace{0.5em}RQ6: Parameter Size and Computational Complexity Analysis.}
Furthermore, we analyze the computational overhead and adaptation efficiency of \modelname{}. The \modelname{} plug-in adds only 345K parameters (2 experts setting) to an AICare~\cite{ma2023aicare} backbone with 2.2M parameters. During training on the MIMIC-IV dataset, we observe that incorporating \modelname{} into AICare introduces approximately one additional second per epoch, primarily due to extra gradient updates and parameters. For the inference time, both with and without TTA require less than one second on the test set (5 forward epochs, batch size of 1024 patients). This modest computational overhead demonstrates an efficient trade-off, as \modelname{} achieves consistent performance improvements across diverse datasets and tasks while maintaining near real-time adaptation capabilities.

\begin{custommdframed}
\textit{Answer to RQ6 (Parameter Size and Computational Complexity Analysis):} \modelname{} demonstrates efficient trade-offs, with a 15.7\% parameter increase (345K added to 2.2M backbone) yielding consistent performance improvements while adding minimal computational overhead (1 second per training epoch, negligible inference delay).
\end{custommdframed}

\section{Discussion}

The performance improvements achieved by \modelname{} highlight several important implications for clinical applications. First, the MoE architecture effectively models the inherent heterogeneity in patient populations by allowing different ``expert'' networks to specialize in distinct patient subgroups. This mimics clinical practice, where medical specialists focus on particular aspects of patient care. Second, the TTA module enables real-time adaptation to new patient data without requiring outcome labels, making \modelname{} particularly suitable for dynamic clinical environments where patient distributions evolve over time.

Our cross-dataset experiments between MIMIC-III and MIMIC-IV demonstrated \modelname{}'s robustness to distributional shifts between different hospital systems and time periods. This capability is especially valuable in healthcare, where clinical protocols, documentation practices, and patient demographics frequently change across institutions and over time. The ablation studies further validated the complementary roles of both the MoE and TTA components, with optimal performance achieved when TTA precedes MoE in the computational flow.

Despite these promising results, several limitations warrant further investigation. First, while \modelname{} improves predictive performance, the clinical interpretability of expert assignments remains unexplored. Understanding which patient characteristics trigger specific expert activations could provide valuable clinical insights and enhance trust in the model's decisions. Second, the behavior of the TTA module when encountering outliers or out-of-distribution samples requires careful evaluation to ensure reliable adaptation in unexpected clinical scenarios.

\section{Conclusion}

In this work, we introduce \modelname{}, a framework that integrates MoE architecture with TTA to address two fundamental challenges in clinical predictive modeling: patient heterogeneity and evolving data distributions. Through extensive experimentation across four real-world EHR datasets, we demonstrate \modelname{}'s ability to consistently improve performance across various clinical prediction tasks and model backbones. In conclusion, \modelname{}'s plug-and-play architecture, consistent performance improvements, and ability to adapt to evolving clinical data make it a promising approach for developing more personalized and robust clinical prediction models.

% References as numbers
\makeatletter
\renewcommand{\@biblabel}[1]{\hfill #1.}
\makeatother

% unstr is used to keep citation order
\bibliographystyle{vancouver}
\bibliography{ref}  

\end{document}